\documentclass[10pt,twocolumn,letterpaper]{article}

\usepackage{cvpr}
\usepackage{times}
\usepackage{epsfig}
\usepackage{graphicx}
\usepackage{amsmath}
\usepackage{amssymb}
\usepackage{bm}
\usepackage[switch]{lineno}
\usepackage{multirow}
\usepackage{xcolor}
\usepackage{colortbl,booktabs}
\usepackage{caption}

\makeatletter
\newif\if@restonecol
\makeatother

\usepackage[linesnumbered,ruled,vlined]{algorithm2e}
\usepackage{algpseudocode}
\usepackage{amsmath}
\usepackage[pagebackref=true,breaklinks=true,letterpaper=true,colorlinks,bookmarks=false]{hyperref}

\cvprfinalcopy 

\def\cvprPaperID{1381} 

\ifcvprfinal\fi
\begin{document}

\title{MWQ: Multiscale Wavelet Quantized Neural Networks}

\author{
Qigong Sun, Yan Ren, Licheng Jiao, Xiufang Li, Fanhua Shang, Fang Liu  \\
{\small Key Laboratory of Intelligent Perception and Image Understanding of Ministry of Education, International Research}\\
{\small Center for Intelligent Perception and Computation, Joint International Research Laboratory of Intelligent Perception}\\
{\small and Computation, School of Artificial Intelligence, Xidian University, Xi'an, Shaanxi Province 710071, China} \\
{\tt\small  xd\_qigongsun@163.com, yanren@stu.xidian.edu.cn, lchjiao@mail.xidian.edu.cn, xfl\_xidian@163.com,} \\
{\tt\small  fhshang@xidian.edu.cn, f63liu@163.com}
}

\maketitle

\begin{abstract}
   Model quantization can reduce the model size and computational latency,
   it has become an essential technique for the deployment of deep neural networks on resource-constrained hardware (e.g., mobile phones and embedded devices).
   The existing quantization methods mainly consider the numerical elements of the weights and activation values, ignoring the relationship between elements.
   The decline of representation ability and information loss usually lead to the performance degradation.
   Inspired by the characteristics of images in the frequency domain, we propose a novel multiscale wavelet quantization (MWQ) method.
   This method decomposes original data into multiscale frequency components by wavelet transform,
   and then quantizes the components of different scales, respectively.
   It exploits the multiscale frequency and spatial information to alleviate the information loss caused by quantization in the spatial domain.
   Because of the flexibility of MWQ,
   we demonstrate three applications (e.g., model compression, quantized network optimization, and information enhancement) on the ImageNet and COCO datasets.
   Experimental results show that our method has stronger representation ability and can play an effective role in quantized neural networks.
\end{abstract}

\vspace{-1 mm}
\section{Introduction}
\vspace{-1 mm}

\begin{figure}[t]
    \setlength{\abovecaptionskip}{0.1cm}
    \setlength{\belowcaptionskip}{-0.3cm}
	\centering
	\includegraphics[width=3in]{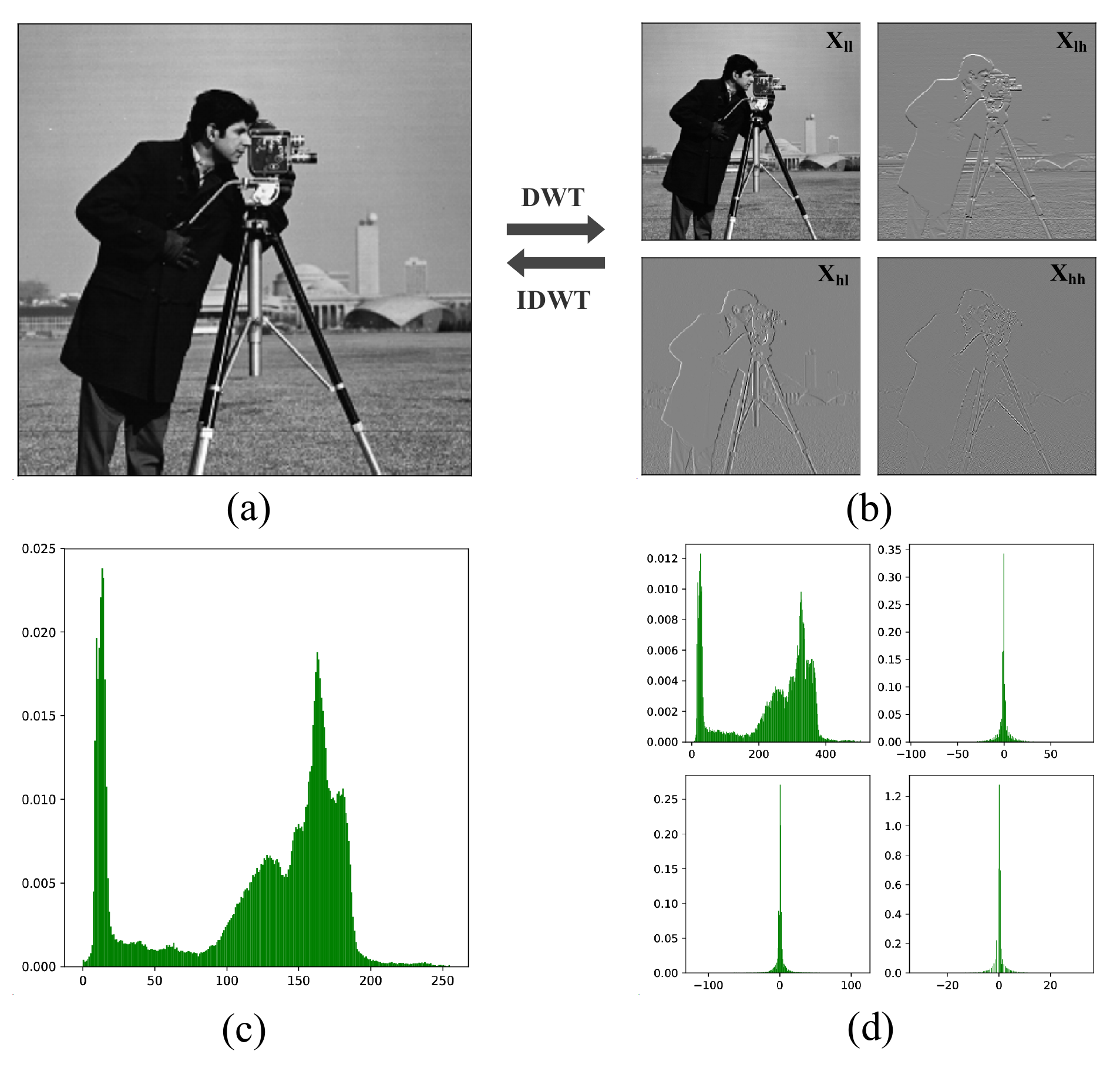}
	\caption{\small{Image visualization and comparison of histograms. 
    (a) Original image.
    (b) High and low frequency components after image decomposition.
    (c) Statistical distribution of original image.
    (d) Statistical distribution of various components in different frequencies after wavelet transform.
    }}
\end{figure}

With the development of artificial intelligence, deep neural networks (DNNs) have achieved remarkable results in many fields, such as image recognition,
natural language processing, and video analysis.
When deeper networks are used to solve various problems, a lot of computing resources and memory are needed urgently.
With the rapid development of chip technologies (e.g., GPU and TPU), the computing frequency and efficiency have been greatly improved.
However, for low-power platforms (e.g., mobile phones, embedded devices, and smart chips) with limited resources,
it is difficult to achieve satisfactory performance (e.g., power dissipation, speed and accuracy) in various industrial applications.
Model quantization is a kind of model compression technology, which transforms the infinite continuous floating-point values into finite discrete fixed-point values.
Recently, some smart chips support low-bit numerical computation, e.g., Apple A12 Bionic \cite{2018apple}, Nvidia Turing GPU \cite{2018Nvidia}, BitFusion \cite{sharma2018bit}, and  BISMO \cite{umuroglu2018bismo}.
Compared with the full-precision model, the quantized model has less parameter storage, lower bandwidth requirements,
faster computing speed, lower energy and memory consumption.
Despite these attractive benefits,
it inevitably leads to the deviation between original data and its quantized value.
Especially when the data is quantized to extreme low-bit, the information loss will cause unstable training and severe performance degradation.


At present, almost all quantization methods directly quantize numerical elements of the weights and activation values, ignoring the relationship between elements.
The decline of representation ability, the loss of information and inadequate training lead to the performance degradation of quantized neural networks.
Spatial information (e.g., corners, edges, textures, and shapes) and multiscale frequency information
play significant roles in image classification, object detection, semantic segmentation, and scene understanding.
Wavelet \cite{mallat1989theory,daubechies1992ten} is a powerful time-frequency analysis tool,
which can be applied to decompose images into various components with different frequencies.
Each component represents different scale information and spatial information.
Fig.\ 1 shows the decomposition of image $\mathbf{X}$ by \emph{Haar} wavelet, and the histograms denote the data distributions before and after discrete wavelet transform (DWT).
The low frequency component $\mathbf{X}_{ll}$ is the down sampling of the original image,
and the high frequency components $\mathbf{X}_{lh}, \mathbf{X}_{hl}$ and $\mathbf{X}_{hh}$ contain more image details (e.g., corners, edges, textures, and shapes) in different directions.
It can be seen from Fig.\ 1 that the different scale components with different distributions have obviously different frequency and spatial information.
These information can not be considered by traditional spatial quantization methods, which lead to information loss and performance degradation.

Inspired by the characteristics of images in the frequency domain,
we propose a novel multiscale wavelet quantization (MWQ) method.
This method uses wavelet transform (e.g., DWT and IDWT) to decompose original data into multiscale components (which contain frequency and spatial information),
and then matches the appropriate quantization parameters (e.g., range and step size) for each component.
By adaptively quantizing each component, it can alleviate the information loss caused by traditional quantization and improve the performance of the quantized model.
The advantages of our method are shown as follows:
\begin{itemize}
    \item \textbf{Stronger representation ability.}
    Assuming that we quantize the weights to $k$-bit, the traditional quantizers can represent up to $2^k$ states.
    For MWQ, quantization is applied in the frequency domain, and each frequency component can be quantized into $2^k$ states.
    After IDWT reconstruction, the number of representation states can be significantly more than $2^k$.
    Therefore, MWQ has stronger representation ability than traditional quantizers, as shown in Fig.\ 6 in Section 4.1.
	\item \textbf{Multiscale quantization.} Wavelet transform can decompose original data to different scales through low pass filters and high pass filters.
    Different scale components contain different information with different distributions, as shown in Fig.\ 1.
    According to the data distribution in each frequency domain, we can clip the data to a more suitable range for rounding to reduce the loss caused by quantization.
	\item \textbf{Spatial relevance.}
    Due to the existence of quantization receptive field, which is defined by the shape of the wavelet base,
    each element after wavelet transform is related to the spatial neighborhoods of original data.
    The proposed method not only considers the numerical value, but also integrates the spatial neighborhood information, as shown in Fig.\ 4.
    \item \textbf{Application flexibility.}
    Compared with the standard Fourier transform, wavelet transform shows diversity based on different wavelet bases.
    The commonly used wavelet bases are \emph{Haar}, \emph{Daubechies}, \emph{Coiflet}, and \emph{Symlets}.
    We can carry out wavelet decomposition with different wavelet base at different level,
    and then implement quantization with various quantizers according to different bit-width requirements.
    Because of the flexibility of MWQ, it can support a variety of applications (e.g., model compression, quantized network optimization, and information enhancement).
    \item \textbf{State-of-the-art results.}
    We apply the proposed techniques on image classification, object detection, and instance segmentation tasks.
    The experimental results show that our techniques are more effective than other counterparts under similar constraints.
\end{itemize}

\vspace{-1 mm}
\section{Related works}

\subsection{Model Quantization}
\vspace{-1 mm}

As one of the typical methods of model compression and acceleration,
model quantization usually quantizes the full-precision parameters to low-bit.
The commonly used quantizers can be categorized into three modalities: uniform quantizers (e.g., PACT \cite{choi2018pact}, Dorefa-Net \cite{zhou2016dorefa}, QIL \cite{jung2019learning} and MBN \cite{sun2019multi}), logarithmic quantizers (e.g., LogQuant \cite{miyashita2016convolutional}, INQ \cite{zhou2017incremental} and ShiftCNN \cite{gudovskiy2017shiftcnn}) and adaptive quantizers (e.g., AdaBits \cite{jin2019adabits} and APoT \cite{li2019additive}).
Under extreme constraints, weights and activation values can also be quantized into binary (e.g., BNN \cite{Courbariaux2016Binarized}, MBN \cite{sun2018efficient} and XNOR-Net \cite{Rastegari2016XNOR}) or ternary (e.g., TWN \cite{li2016ternary} and TTQ \cite{zhu2016trained}), which can be computed by bitwise operations (e.g., xnor, and bitcount).
Some studies \cite{wang2019haq, wu2018mixed,cai2020rethinking,sun2021effective} show that mixed-precision quantization can achieve better performance according to the sensitivities of different layers.
It is hardly to access the accurate gradients for discrete quantization, and the commonly used method is to resort to appropriate approximation.
Straight through estimation (STE) \cite{bengio2013estimating,feng2020convolutional} uses the nonzero gradient to approximate the function gradient, which is not-differentiable or whose derivative is zero.
When the values are quantized to extremely low-bit, STE greatly harms the performance of quantized models \cite{li2017training,mckinstry2018discovering}.
DSQ \cite{gong2019differentiable} employs a series of hyperbolic tangent functions to gradually approach the staircase function for low-bit quantization, which makes
the forward and backward process more consistent and stable in the training.
However, it can not be used to approximate the nonuniform quantization.

\begin{figure*}[!htb]
    \setlength{\abovecaptionskip}{0.1cm}
    \setlength{\belowcaptionskip}{-0.4cm}
	\centering
	\includegraphics[width=5.6in]{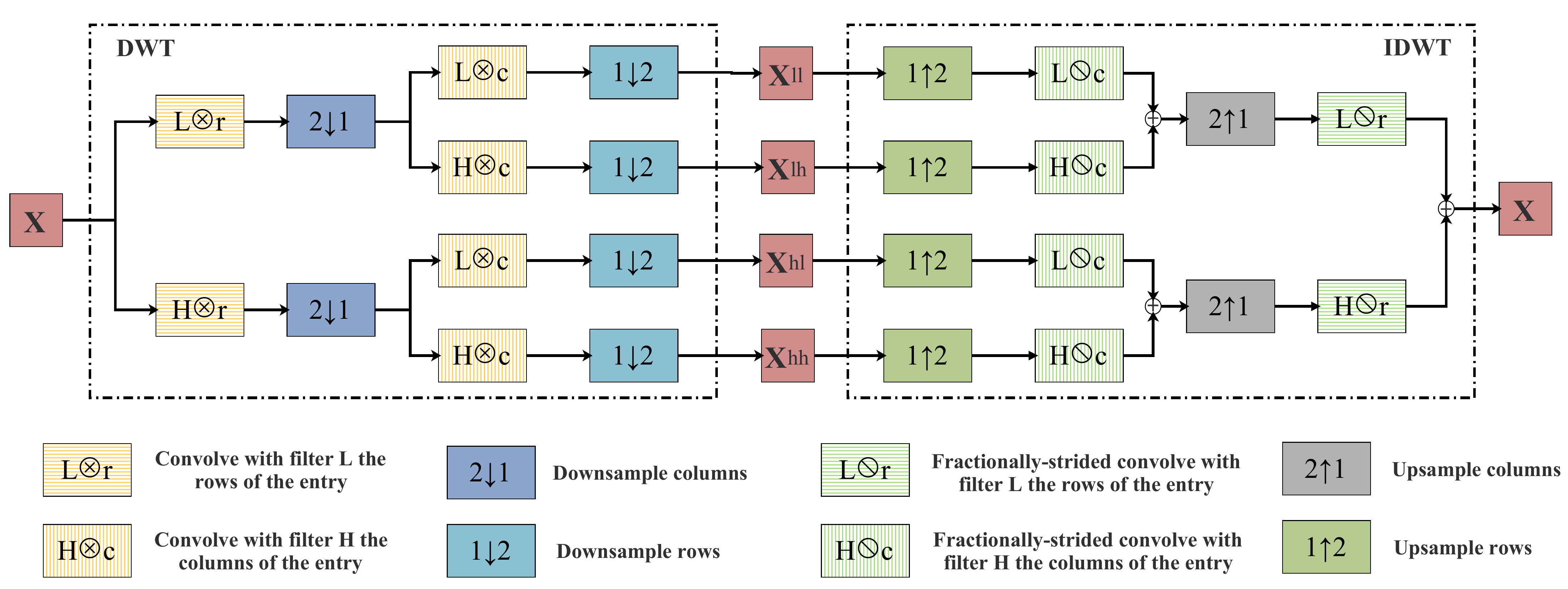}
	\caption{\small{Illustration of DWT and IDWT for image data. The low-frequency component $\mathbf{X}_{ll}$ contains the main information of original image and
    the high-frequency components $\mathbf{X}_{lh}, \mathbf{X}_{hl}$ and $\mathbf{X}_{hh}$ denote the vertical, horizontal and diagonal components of image.
    }}
\end{figure*}

\vspace{-1 mm}
\subsection{Wavelets}
\vspace{-1 mm}

Wavelet is derived from multi-resolution analysis, and it expresses a function as a series of successive approximation components, and
each component represents a different resolution.
The commonly used wavelets include orthogonal wavelets, biorthogonal wavelets, multiwavelets, ridgelet, curvelets, bandelets, and contourlets, etc.
This technique is often used for function approximation \cite{zhang1992wavelet} and signal processing \cite{mallat1996wavelet,szu1992neural}.
The discrete wavelet transform (DWT) can be used to decompose an image into different levels of frequency interpretations,
and the inverse discrete wavelet transform (IDWT) can reconstruct original image by using the multi-frequency components.
In image processing, wavelet transform is often used as a tool for content information analysis \cite{mallat1989theory}.
With the development of DNNs, wavelet transform has several attempts to combine the classical signal processing and deep learning methods,
such as image denoising \cite{kang2018deep,liu2020densely,wang2020multi}, super resolution \cite{huang2017wavelet,liu2018multi}, classification \cite{de2020multi,li2020wavelet,liu2020c}, segmentation \cite{li2020wavesnet}, facial aging \cite{liu2019attribute}, style transfer \cite{yoo2019photorealistic},  remote sensing image processing \cite{duan2017sar}, etc.
It is often used as the tool of data preprocessing, post-processing, feature extraction, and sampling operators in DNNs \cite{huang2017wavelet,liu2019attribute,savareh2019wavelet,williams2018wavelet,liu2018multi,li2020complex}.
\cite{li2020wavelet} utilizes DWT to replace max-pooling, strided-convolution, and average-pooling to suppress the noise effect.
Multi-level wavelet CNN \cite{liu2018multi} use DWT to concatenate the frequency and position information of feature mapping which preserve texture details.
\cite{liu2019attribute} incorporates a wavelet packet transform module to improve the visual fidelity of generated images by capturing age-related texture details at multiple scales in the frequency space.

\vspace{-1 mm}
\section{Approach}
\vspace{-1 mm}

In this paper, we innovatively view the model quantization from the perspective of the frequency domain.
Through wavelet transform, multiscale frequency and spatial information
are considered to alleviate the information loss and performance degradation caused by quantization.
In this section, we first review the process of DWT and IDWT operations, and discuss the feasibility in DNNs.
Then we analyze the problem of traditional quantizers and introduce our method in more detail.


\vspace{-1 mm}
\subsection{DWT and IDWT Operations}
\vspace{-1 mm}

Similar to Fourier series that use the combination of sine functions to represent a discrete signal, wavelet can decompose a signal in different frequency domains.
DWT can decompose a signal into multiscale components with different frequencies.
After multi-level DWT, the original signal $\mathbf{s}$ can be mathematically represented by the summation of approximated version $\mathbf{s}^l$ and detailed versions $\mathbf{s}^h$ as defined below.
\setlength{\abovedisplayskip}{3pt}
\setlength{\belowdisplayskip}{3pt}
\begin{eqnarray}
\mathbf{s} = \mathbf{s}^h_1 \oplus \mathbf{s}^l_1 = \mathbf{s}^h_1 \oplus \mathbf{s}^h_2 \oplus \cdots \oplus  \mathbf{s}^h_n \oplus  \mathbf{s}^l_n,
\end{eqnarray}
where $n$ represents the decomposition level.
Suppose $\mathbf{s}=\{s[n]\}_{n \in [1,N]}$ is a 1D discrete signal.
The DWT of this signal is calculated by passing it through a series of filters (e.g., low pass filter $\mathbf{g}=\{g[k] \}$ and high pass filter $\mathbf{h}=\{h[k] \}$).
The first level low frequency component $\mathbf{s}^l_1$ and high frequency component $\mathbf{s}^h_1$ can be expressed as follows:
\setlength{\abovedisplayskip}{3pt}
\setlength{\belowdisplayskip}{3pt}
\begin{eqnarray}
{s}^l_1[k] = \sum _j s[j]g[j-2k], \\
{s}^h_1[k] = \sum _j s[j]h[j-2k].
\end{eqnarray}
IDWT can use the decomposed components to reconstruct the original signal as follows:
\setlength{\abovedisplayskip}{3pt}
\setlength{\belowdisplayskip}{3pt}
\begin{eqnarray}
{s}[j] = \sum _k (s^l_1[k]g[j-2k] + s^h_1[k]h[j-2k]).
\end{eqnarray}

In image processing, the original image $\mathbf{X}$ can be decomposed into one low frequency component $\mathbf{X}_{ll}$ and three high frequency components $\mathbf{X}_{lh}, \mathbf{X}_{hl}$ and $\mathbf{X}_{hh}$ by single-level DWT, where the low frequency component denotes the approximation coefficient and the high frequency components contain more image details (e.g., edges, shapes and textures).
For a given image $\mathbf{X} \in \mathbb{R}^{m \times n}$, the DWT decomposition can be expressed as follows:
\setlength{\abovedisplayskip}{3pt}
\setlength{\belowdisplayskip}{3pt}
\begin{small}
\begin{eqnarray}
\mathbf{X}_{c_0c_1}=(1\downarrow2)(\mathbf{f}_{c_1} \otimes _c (2\downarrow1)(\mathbf{f}_{c_0} \otimes _r \mathbf{X})), c_0,c_1 \in \{l,h\},
\end{eqnarray}
\end{small}
where $\otimes _r$ and $\otimes _c$ denote the convolution for the rows and columns of the entry,
$(2\downarrow1)$ and $(1\downarrow2)$ denote the downsample columns and rows, respectively.
$\mathbf{f}_{l}$ and $\mathbf{f}_{h}$ represent the low pass filter and high pass filter, which are determined by the wavelet base (e.g., \emph{Haar} and \emph{Daubechies}).
Similar to DWT, various frequency components can also be used to reconstruct the original image as follows:
\setlength{\abovedisplayskip}{3pt}
\setlength{\belowdisplayskip}{3pt}
\begin{small}
\begin{eqnarray}
\mathbf{X}=\sum_{c_1}\mathbf{f}_{c_1}\oslash _r (2\uparrow1)(\sum_{c_0}\mathbf{f}_{c_0}\oslash _c (1\uparrow2) \mathbf{X}_{c_0c_1}),
\end{eqnarray}
\end{small}
where $c_0,c_1 \in \{l,h\}$, $\oslash _r$ and $\oslash _c$ denote the deconvolution for the rows and columns of the entry,
$(2\uparrow1)$ and $(1\uparrow2)$ denote the upsample columns and rows, respectively.

Fig.\ 2 describes the basic decomposition and reconstruction steps for an image.
In order to facilitate the application of wavelet transformation in DNNs,
we use convolution and fractionally-strided convolution to implement DWT and IDWT operations in PyTorch.
The low pass filter and high pass filter can be defined as convolution kernels.
The convolution direction is controlled by the shape of convolution filter, and the upsample and downsample are controlled by stride.
Therefore, DWT and IDWT can be used to participate in the forward and backward propagations of DNNs.

\vspace{-1 mm}
\subsection{Traditional Quantizer}
\vspace{-2 mm}

\begin{figure}[!h]
    \setlength{\abovecaptionskip}{0.1cm}
    \setlength{\belowcaptionskip}{-0.3cm}
	\centering
	\includegraphics[width=3.1in]{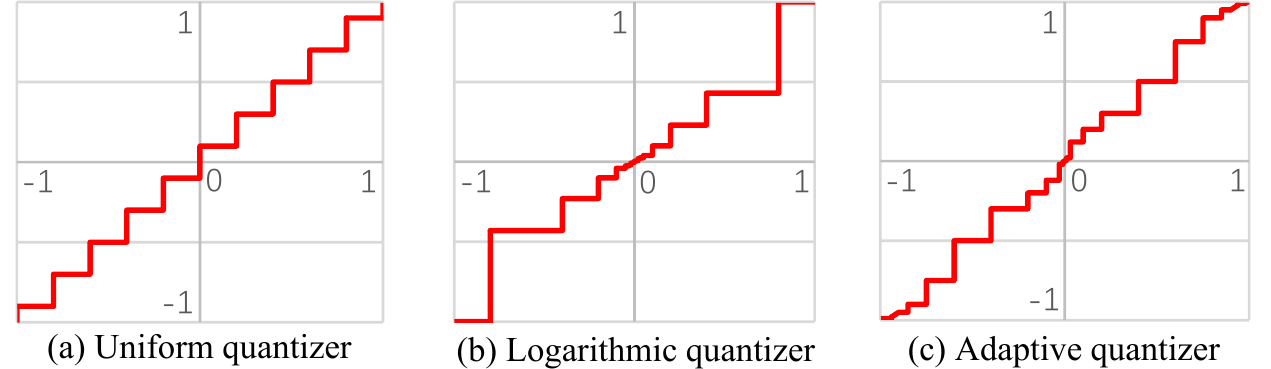}
	\caption{\small{Schematic diagram of the three quantizers.
    }}
\end{figure}

Model quantization is an effective model compression technique.
For example, if we quantize the weights to 4-bit, we can achieve $8 \times$ of the model compression ratio.
Suppose we use $\mathbf{X}^i$ and $\mathbf{X}^{i+1}$ to represent the input data and output data of the $i$-th layer.
$\mathbf{W}^{i}$ denotes the weights, and $\mathcal{F}(\,)$ denotes the calculation operations (e.g., full connection or convolution).
The calculation between two variables can be shown as follows:
\setlength{\abovedisplayskip}{3pt}
\setlength{\belowdisplayskip}{3pt}
\begin{eqnarray}
\mathbf{X}^{i+1}=\mathcal{F}(\mathbf{\hat{X}}^{i}_n,\mathbf{\hat{W}}^{i}_m) = \mathcal{F}(\mathcal{Q}(\mathbf{X}^i,n),\mathcal{Q}(\mathbf{{W}}^{i},m)),
\end{eqnarray}
where $\mathcal{Q}(\,)$ denotes the quantizer, and $n$ and $m$ represent the bit-widths of input data and weights.
The calculation of quantized variables ($\mathbf{\hat{X}}^{i}_n$ and $\mathbf{\hat{W}}^{i}_m$) can be accelerated by a special accelerator (e.g., BitFusion \cite{sharma2018bit} and  BISMO \cite{umuroglu2018bismo}).
Fig.\ 3 shows the schematic diagram of uniform quantizer, logarithmic quantizer, and adaptive quantizer.
Existing quantizers learn the range and step size of quantization by data distribution.
They only focus on the numerical value and ignore the spatial information of the data.
Therefore, it is easy to cause the loss of spatial information (e.g., edges, textures, and shapes) and severe performance degradation.

\vspace{-1 mm}
\subsection{Multiscale Wavelet Quantization}
\vspace{-1 mm}

In order to alleviate the information loss caused by quantization,
we propose a novel multiscale wavelet quantization method based on the characteristics of wavelet transform.
We use the DWT operation to decompose the original data into its multiscale frequency components.
The decomposed components represent the mapping of the original data at different scales.
Different components contain different contents and and have different distributions.
Therefore, we can use the traditional quantizer (e.g., Uniform and APoT) to quantize each components separately to reduce the information loss.
Finally, we reconstruct the quantized components by the IDWT operation to serve as network computing.
The main process can be expressed as follows:
\setlength{\abovedisplayskip}{3pt}
\setlength{\belowdisplayskip}{3pt}
\begin{eqnarray}
\mathbf{{X}}^{i}_{ll},\mathbf{{X}}^{i}_{lh},\mathbf{{X}}^{i}_{hl},\mathbf{{X}}^{i}_{hh} = \mathcal{DWT}(\mathbf{{X}}^{i}, {name}, {J}), \\
\mathbf{\hat{X}}^{i}_{c_0,c_1} = \mathcal{Q}(\mathbf{{X}}^{i}_{c_0,c_1},n), c_0,c_1 \in \{l,h\}\\
\mathbf{\hat{X}}^{i} = \mathcal{IDWT}(\mathbf{\hat{X}}^{i}_{ll},\mathbf{\hat{X}}^{i}_{lh},\mathbf{\hat{X}}^{i}_{hl},\mathbf{\hat{X}}^{i}_{hh}, {name}, {J})
\end{eqnarray}
where $\mathcal{DWT}(\,)$ and $\mathcal{IDWT}(\,)$ represent the DWT and IDWT operations defined in Section 3.1,
${name}$ denotes the wavelet base (e.g., \emph{Haar}, \emph{Daubechies}, \emph{Coiflet}, and \emph{Symlets}), and $J$ denotes the level of wavelet decomposition.
$\mathcal{Q}(\,)$ denotes the quantizer, in which the quantization range and step size are different for different components.
$\mathbf{\hat{X}}^{i}_{ll}$, $\mathbf{\hat{X}}^{i}_{lh}$, $\mathbf{\hat{X}}^{i}_{hl}$ and $\mathbf{\hat{X}}^{i}_{hh}$ are the quantized multiscale frequency components.

\vspace{-2 mm}
\begin{figure}[!htb]
    \setlength{\abovecaptionskip}{0.1cm}
    \setlength{\belowcaptionskip}{-0.3cm}
	\centering
	\includegraphics[width=3.0in]{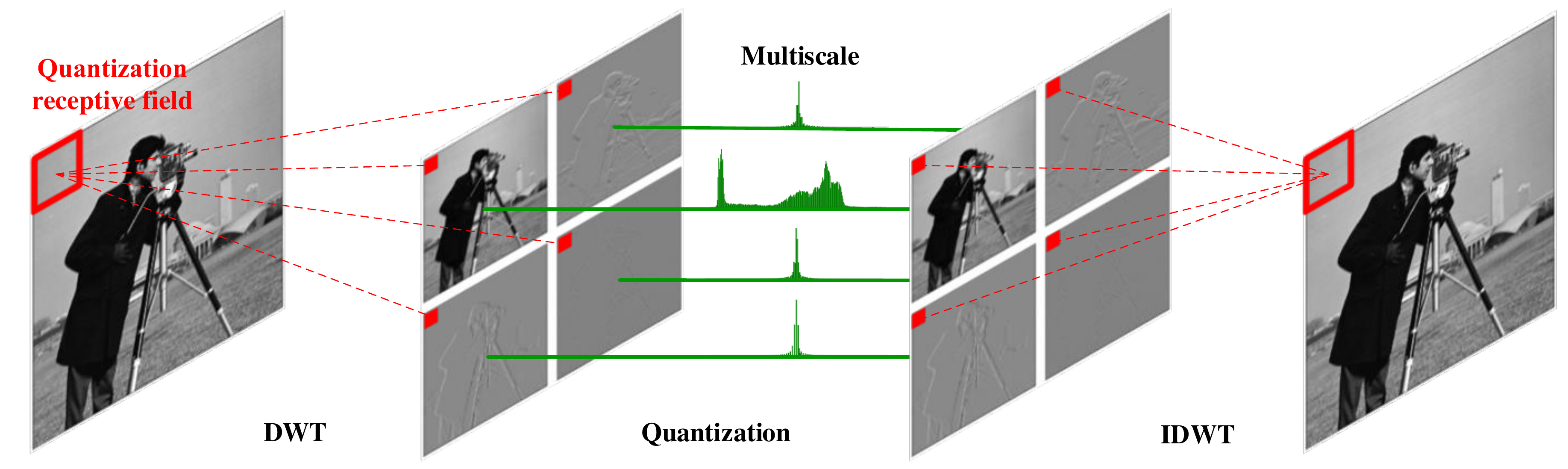}
	\caption{\small{Schematic diagram of multiscale wavelet quantization.
    }}
\end{figure}

In addition to considering the multiscale information in each frequency domain, our method also takes into account the spatial neighborhood information.
Fig.\ 4 shows the schematic diagram of MWQ.
Obviously, the low frequency component is the down sampling of the original data, which has a similar distribution as the original data.
High frequency components mainly present the details of the original data.
As can be seen from Fig.\ 4, each element ({red unit}) of the frequency component
is derived from the spatial neighborhood ({red block}) of the original data.
The block in the spatial domain is called \textbf{quantization receptive field}, which is defined by the shape of the wavelet filter.
For example, the quantization receptive fields of \emph{Haar} and \emph{Daubechies} are 2 and 4, respectively.
Therefore, MWQ considers the correlation of spatial information and preserves the spatial information as much as possible.

\vspace{-1 mm}
\section{Applications and Experiments}
\vspace{-1 mm}

MWQ can make use of the decomposed multiscale frequency information and spatial neighborhood information
to alleviate the performance degradation caused by the quantization in the spatial domain.
In order to verify the effectiveness of our proposed quantization method, we apply it to image classification, object detection and instance segmentation tasks.
The first experiment is model compression, which only focuses on the quantization of weights.
MWQ is applied on the weights, and the activation values are in full-precision form or quantized by traditional quantizers.
Then we verify the effectiveness of our method in the optimization of the traditional quantized networks.
Finally, we use the multiscale features of wavelet decomposition to enhance the high frequency component information,
and verify the importance of spatial information (e.g., edges, textures, and shapes).
The implementation details and experimental results are shown below.


\textbf{Quantizer}:
We use the hardware friendly uniform quantization function as the quantizer for the following experiments.
The $m$-bit quantization of weights $\mathbf{W}^i$ can be formulated as follows:
\setlength{\abovedisplayskip}{4pt}
\setlength{\belowdisplayskip}{4pt}
\begin{eqnarray}
\mathcal{Q}(\mathbf{W}^i, m) = \textrm{round}(\textrm{clamp}(\mathbf{W}^i/s^i,-1, 1)*S)*d^i, \\
s.t.\quad S = 2^{m-1}-1, \quad d^i = s^i/S, \quad\quad\quad\quad\quad\:\: \nonumber
\end{eqnarray}
where the clamp function is used to truncate all values into the range of $[-1, 1]$, $s^i$ is a learned parameter of the $i$-th layer,
and $d^i$ denotes the scaling factor.
Because the activation values after the ReLU function are all non-negative, the quantization of activation values will be truncated into the range of $[0, 1]$.

\vspace{-1 mm}
\subsection{Model Compression}
\vspace{-1 mm}

Quantization compresses the original network by reducing the bits of weights \cite{cheng2017survey}.
Under extreme constraints, the model can be quantized to binary \{-1, +1\} to achieve nearly a $32 \times$ model compression ratio \cite{Courbariaux2016Binarized, Rastegari2016XNOR}.
The common way is to train a quantized model in the spatial domain for storage, transmission and calculation, as shown in Fig.\ 5 (a).
A novel idea is proposed, that is, the weights are decomposed into the frequency domain for quantization, storage and transmission.
Thus, more information is preserved.
In this application, we focus on the quantization of the weights for the classification on ImageNet dataset,
and all the activation values are in full-precision form or quantized by the traditional quantizers.
Fig.\ 5 (b) shows the schematic diagram of our method for model compression,
where $\mathbf{{W}}^{i}_{ll}$, $\mathbf{{W}}^{i}_{lh}$, $\mathbf{{W}}^{i}_{hl}$ and $\mathbf{{W}}^{i}_{hh}$ are the multiscale frequency components decomposed by DWT operation.
$\mathbf{\hat{W}}^{i}_{ll}$, $\mathbf{\hat{W}}^{i}_{lh}$, $\mathbf{\hat{W}}^{i}_{hl}$ and $\mathbf{\hat{W}}^{i}_{hh}$ are the quantized multiscale frequency components.
We can also perform multi-level decomposition and reconstruction for more detailed quantization, as shown by the dotted line in Fig.\ 5 (b).


\begin{figure}[t]
    \setlength{\abovecaptionskip}{0.1cm}
    \setlength{\belowcaptionskip}{-0.4cm}
	\centering
	\includegraphics[width=3.0in]{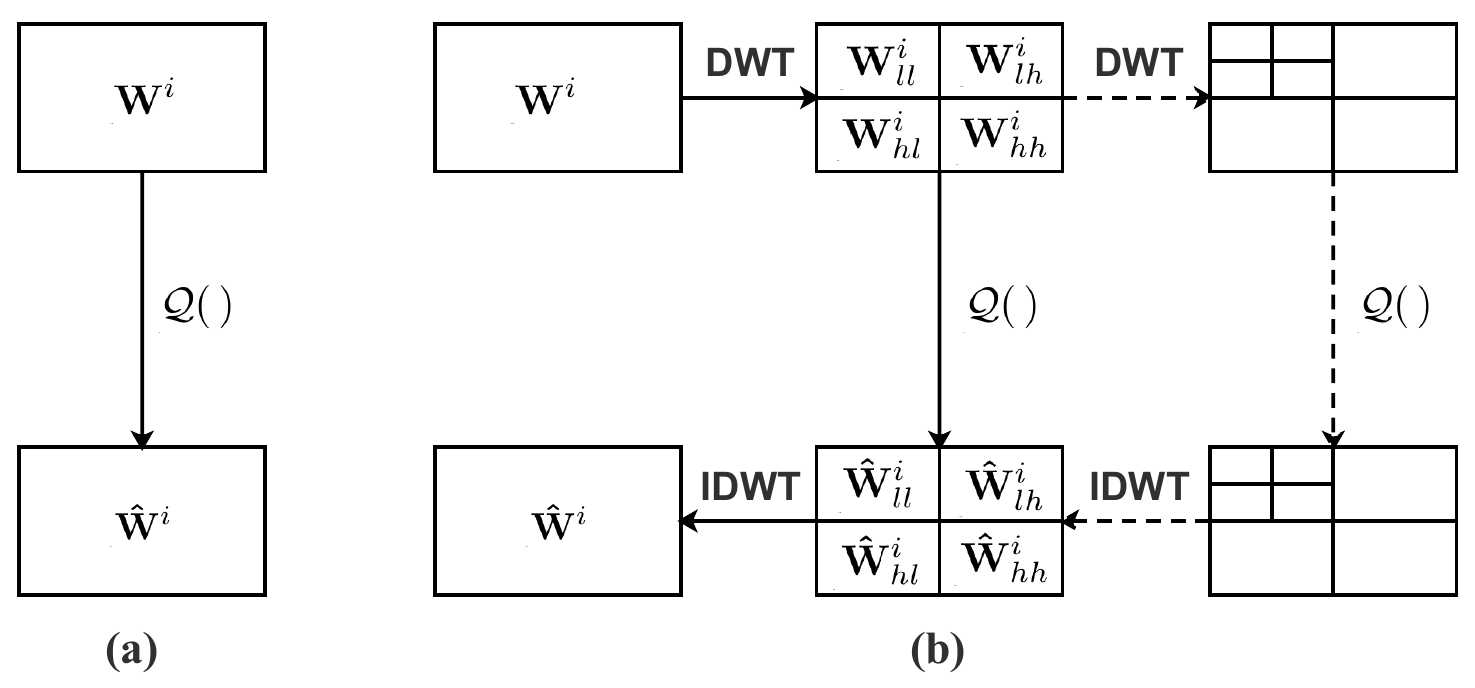}
	\caption{\small{(a) Traditional quantization. (b) Wavelet quantization.
    }}
\end{figure}

\textbf{Experimental results:}
The ImageNet dataset (ILSVRC2012) consists of images of 1K categories, and has over 1.2M images in the training dataset and 50K images in the validation dataset.
It is a common large-scale dataset for image classification tasks.
In order to verify the effectiveness of our method on large-scale datasets and deep networks,
we implement experiments with ResNet-18 and ResNet-50 on this dataset.
In the experiments, the activation values are in full-precision form or are quantized to 4-bit.
Following previous methods \cite{zhuang2020training,zhou2016dorefa}, we quantize the first convolutional layer and the last fully-connected layer to $8$-bit.
We apply the pre-trained full-precision model to initialize the quantized model.
Network parameters $\mathbf{W}$ are updated 50 epochs by SGD, the initial learning rate is $1 \times 10^{-2}$.
The learning rate is decayed by a factor of 10 at epochs 10, 25, and 40, respectively.
For ImageNet, the batch size for all the networks is set to $1024$.
Here, we compare the performance of our method with those of state-of-the-art methods,
such as TWN \cite{li2016ternary}, TTQ \cite{zhu2016trained}, LQ-Nets \cite{zhang2018lq}, QIL \cite{jung2019learning}, DSQ \cite{gong2019differentiable}, HAQ \cite{wang2019haq}, and HAWQ \cite{dong2019hawq}.
HAQ and HAWQ are typical mixed-precision quantization methods.

\begin{table}[!t]
    \setlength{\abovecaptionskip}{0.0cm}
    \setlength{\belowcaptionskip}{-0.4cm}
    \footnotesize
	\centering
	\caption{Accuracy comparisons of ResNet-18 and ResNet-50 on ImageNet.
    `M' refers to mixed-precision quantization.
    }
	\setlength{\tabcolsep}{1.4mm}{
		\renewcommand\arraystretch{1.15}
		\begin{tabular}{ccccccc}
			\hline
			Models & Methods    & W-Bits    & A-Bits    & Top-1    & W-Comp  \\
			\hline
            \multirow{24}*{ResNet-18}
			&FP        & 32    & 32  & 70.20     & 1.00           \\ \cline{2-6}
			&TWN       & 2     & 32   & 65.30    & 16.00        \\
			&TTQ       & 2     & 32   & 66.60    & 16.00         \\
			&LQ-Nets   & 2     & 32   & 68.00    & 16.00         \\ \cline{2-6}
			&\multirow{4}*{MWQ [Haar]}
			   & [2,2,2,2], $J$=1      & 32   & \textbf{68.79}    & 16.00            \\
			&  & [3,2,2,1], $J$=1      & 32   & 68.27    & 16.00        \\
            &  & [4,2,1,1], $J$=1      & 32   & 67.41    & 16.00       \\
            &  & [5,1,1,1], $J$=1      & 32   & 67.00    & 16.00       \\\cline{2-6}
			&LQ-Nets   & 3     & 32   & 69.30    & 10.67           \\
			&QIL       & 3     & 32   & 69.90    & 10.67          \\ \cline{2-6}
			&\multirow{4}*{MWQ [Haar]}
			   & [3,3,3,3], $J$=1      & 32   & \textbf{70.04}    & 10.67            \\
			&  & [4,3,3,2], $J$=1      & 32   & 69.80    & 10.67        \\
            &  & [5,3,2,2], $J$=1      & 32   & 69.74    & 10.67       \\
            &  & [6,2,2,2], $J$=1      & 32   & 69.43    & 10.67       \\\cline{2-6}
			&PACT       & 4     & 4   & 69.20    & 8.00         \\
			&LQ-Nets    & 4     & 4   & 69.30    & 8.00            \\
			&DSQ        & 4     & 4   & 69.56    & 8.00           \\
			&QIL        & 4     & 4   & 70.10    & 8.00           \\ \cline{2-6}
            &MWQ [Haar] & [4,4,4,4], $J$=1      & 4   & 71.01    & 8.00            \\
            &MWQ [db2] & [4,4,4,4], $J$=1       & 4   & 71.02    & 8.00            \\
			&MWQ [sym2] & [4,4,4,4], $J$=1      & 4   & \textbf{71.04}    & 8.00            \\
			&MWQ [coif2] & [4,4,4,4], $J$=1     & 4   & 70.92    & 8.00            \\
			\hline
            \multirow{9}*{ResNet-50}
			&FP   & 32    & 32  & 77.15    & 1.00       \\ \cline{2-6}
			&LQ-Nets      & 4     & 4   & 75.10    & 8.00      \\
			&HAQ        & M     & M   & 75.48    & $\sim$ 8.00      \\
			&HAWQ       & M     & M   & 75.30    & $\sim$ 8.00      \\ \cline{2-6}
            &MWQ [Haar] & [4,4,4,4], $J$=1      & 4   & 76.33     & 8.00            \\
            &MWQ [db2] & [4,4,4,4], $J$=1      & 4   & 76.32    & 8.00            \\
			&MWQ [sym2] & [4,4,4,4], $J$=1     & 4   & 76.23    & 8.00            \\ 
			&MWQ [coif2] & [4,4,4,4], $J$=1    & 4   & {76.36}    & 8.00            \\
			&MWQ [coif2] & [4,4,4,4], $J$=2    & 4   & \textbf{76.37}    & 8.00            \\
			\hline
		\end{tabular}
	}
	\label{tab:Resnettable}
\vspace{-5 mm}
\end{table}

For each compared method, we report its weight bit, activation value bit, Top-1 accuracy, and weight compression rate.
Table 1 shows all the experimental results.
We use `MWQ [Haar] - [6,2,2,2], $J$=1  - 32' as an example to explain the result.
`MWQ' denotes multiscale wavelet quantization and `Haar' denotes the wavelet base for decomposition.
`[6,2,2,2], $J$=1' denotes that the weights are decomposed by single-level wavelet,
and the low frequency component is quantized to 6-bit and other high frequency components are quantized to 2-bit.
`32' refers to the activation values being in full-precision form.
We quantize the weights based on multiple wavelet bases (e.g., \emph{Haar}, \emph{Daubechies}, \emph{Symlets} and \emph{Coiflets}).
In Table 1, `bd2' represents the orthogonal \emph{Daubechies} wavelet with approximation order 2.
From Table 1, we can see that our method can obtain better classification accuracies than other methods.
Especially when the activation values are quantized to 4-bit, the accuracies of our method are obviously higher than their full-precision counterparts.
When we use different bit-widths to quantize the frequency components of the weights, we find that the experimental results have obvious differences.
The high frequency components of the weights also have a great contribution to the quantization effect.
When we decompose weights with different wavelets, the experimental results are almost equivalent.
Therefore, the decomposition based on multiple wavelet bases is an effective method for model compression.

We take the weights of `layer2.0.conv1' of ResNet-18 for the classification on ImageNet as an
example to analyze the numerical distribution of MWQ, the weight statistics histograms are shown in Fig.\ 6.
(a) is the statistics histograms of original weights, (b) is the statistics histograms of quantized 4-bit weights,
and (c) is the statistics histograms of quantized 4-bit weights by our method.
Obviously, we can see that the weights quantized by our method have more representation states.
Therefore, MWQ has stronger representation ability and can achieve better performance under the same model compression ratio.

\begin{figure}[!htb]
    \setlength{\abovecaptionskip}{0.1cm}
    \setlength{\belowcaptionskip}{-0.4cm}
	\centering
	\includegraphics[width=2.8in]{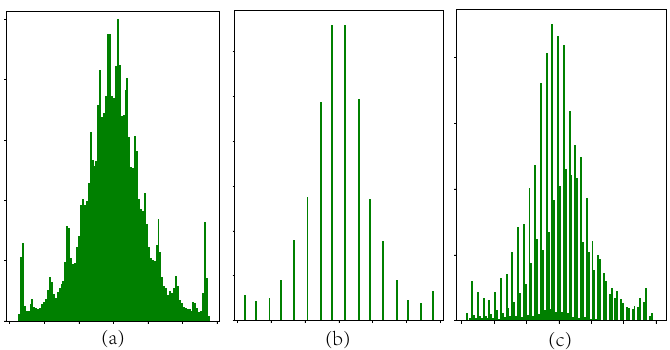}
	\caption{\small{Comparison of weight statistics histograms of `layer2.0.conv1' of ResNet-18 before and after multiscale wavelet quantization.
    (a) Statistics histogram of original weights.
    (b) Statistics histogram of original weights after 4-bit quantization.
    (c) Statistics histogram of data after IDWT reconstruction.
    }}
\end{figure}

\vspace{-1 mm}
\subsection{Quantized Network Optimization}
\vspace{-1 mm}

Quantized neural network is a typical discrete neural network.
The derivative of the quantization function is not defined, and thus traditional gradient optimization methods are not applicable.
STE \cite{bengio2013estimating} updates gradients by predefining a fixed derivative,
which is widely applied in \cite{lin2017towards,zhang2018lq,sun2019multi}.
Experiments and analysis in \cite{li2017training,mckinstry2018discovering} show that the gradient error caused by
quantization and STE greatly harms the accuracy of quantized models when models are quantized to low-bit.
DSQ \cite{gong2019differentiable} employs a series of hyperbolic tangent functions to gradually approach the staircase function for low-bit quantization.
However, it can only be used for the optimization of uniform quantization.
Progressive quantization (PQ) \cite{zhuang2018towards} uses the high-bit model to initialize the low-bit weights, which can speed up the convergence and improve accuracy.

\begin{figure*}[!htb]
    \setlength{\abovecaptionskip}{0.2cm}
    \setlength{\belowcaptionskip}{-0.3cm}
	\centering
	\includegraphics[width=6.2in]{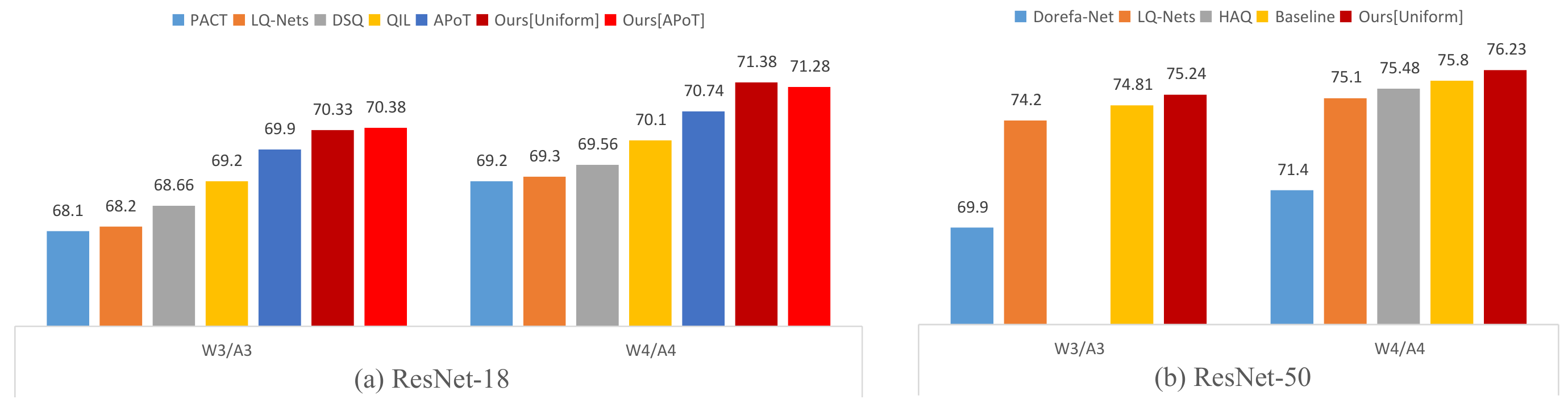}
	\caption{\small{The experimental results of these quantized neural networks based on ResNet-18 and ResNet-50.
    }}
\end{figure*}

From the Application 1, we can see that MWQ has stronger representation ability,
and provides better performance for quantized networks.
Therefore, we use the ideology of PQ for reference and introduce MWQ into the optimization of traditional quantized neural networks.
In this application, different from the multistage training pipelines in PQ, we only consider two types of bitwidth precision for weights quantization.
In other words, we first train an MWQ model (just like Application 1), and then use the trained model to initialize the target quantized network.
As an orthogonal method, our method can support multiple quantizers (e.g., the uniform quantizer and adaptive quantizer).
The iterative procedure is shown in Algorithm 1.

\begin{algorithm}
  \caption{Quantized Network Optimization based on Multiscale Wavelet Quantization}
  \KwIn{The training dataset $\{(\mathbf{X}_i, \mathbf{y}_i)\}^N_{i=1}$;
		Initializing parameters by a pre-trained model. \\}
  \KwOut{A model with weights and activations being quantized into $k$-bit. \\}
  \textbf{Stage 1}: Training multiscale wavelet quantized model: \\
  \For{$epoch=1,\ldots,L$}
  {
    \For{$t=1,\ldots,T$}
    {
    Quantizing the activations into $k$-bit; \\
    Using Eqs.\ $(8)\sim(10)$ to quantize the weights into $k$-bit; \\
    Updating weights; \\
    }
  }
  \textbf{Stage 2}: Fine-tuning quantized model: \\
  Initializing parameters using the trained $k$-bit multiscale wavelet quantized model from \textbf{Stage 1}; \\
  \For{$epoch=1,\ldots,L$}
  {
    \For{$t=1,\ldots,T$}
    {
    Quantizing the activations into $k$-bit; \\
    Quantizing the weights into $k$-bit; \\
    Updating weights; \\
    }
  }
\end{algorithm}

\textbf{Experimental results:}
To verify the effectiveness of our proposed optimization method, we still use the ImageNet dataset to optimize ResNet-18 and ResNet-50 models.
We take single-level decomposition by the \emph{Haar} wavelet for MWQ, and
the same optimization strategies as in Application 1 are adopted to train a multiscale wavelet quantized model.
After 50 epochs, the $k$-bit multiscale wavelet quantized model will be used to initialize the parameters of Stage 2.
In Stage 2, the initial learning rate is $1 \times 10^{-3}$ and decayed by a factor of 10 at epochs 20 and 40.

Fig.\ 7 shows the comparable experimental results on ResNet-18 and ResNet-50.
For ResNet-18, we have carried out experiments on two kinds of quantizers (e.g., Uniform and APoT).
The two-stage optimization method with MWQ has obvious advantages, and its accuracies are over 70\% in 3 bits and 71\% in 4 bits.
For ResNet-50, we use the trained model by PQ as the baseline (yellow).
Compared with baseline, our method is 0.43\% higher in 3-bit and 4-bit.

\vspace{-1 mm}
\subsection{Information Enhancement}
\vspace{-1 mm}

In image processing tasks, features (e.g., edges, textures, and shapes) play important roles in image classification, object detection, semantic segmentation, and scene understanding.
Therefore, it is especially important for image understanding to detect the edge from the small outline of the structure to the boundary of the large visual object.
DNNs use multiple hidden layers to achieve feature extraction of images automatically.
Due to the lack of supervision on the details of specific regions, the spatial distribution of semantic features will be confused.
This phenomenon greatly weakens the representation ability of features and brings difficulties to the construction of hierarchical understanding.
SEAnet \cite{chensupervised} utilizes edge attention to highlight the object and suppress background noise, and
SPNet \cite{hou2020strip} proposes strip pooling to aggregate global and local contexts for scene parsing.

As a tool of multiscale frequency analysis, wavelet can be used for image decomposition.
The low frequency component stores the specific information of the image,
and the high frequency components store significant information (e.g., edges, textures, and shapes) in different directions.
In order to verify the effectiveness of high frequency information and alleviate the loss of spatial information in the process of feature extraction,
we consider to decompose the feature maps by wavelet transform and enhance the high frequency information.
We introduce scale factors $\alpha$ into the high frequency components.
Fig.\ 8 illustrates the main process, where $\mathbf{X}$ and $\hat{\mathbf{X}}$ denote the original feature map and enhanced feature map, respectively.

\vspace{-1 mm}
\begin{figure}[!htb]
    \setlength{\abovecaptionskip}{0.2cm}
    \setlength{\belowcaptionskip}{-0.3cm}
	\centering
	\includegraphics[width=3.2in]{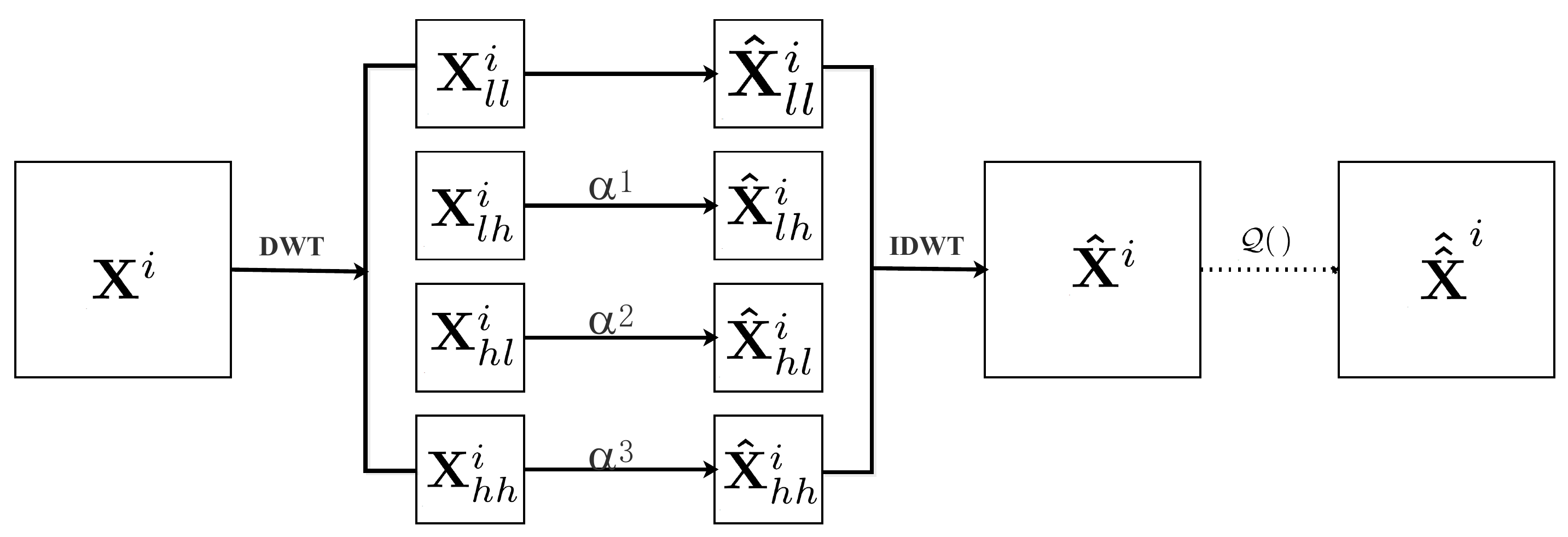}
	\caption{\small{Schematic diagram of information enhancement.
    }}
\end{figure}

\begin{table*}[!t]
    \setlength{\abovecaptionskip}{0.2cm}
    \setlength{\belowcaptionskip}{-0.4cm}
    \footnotesize
    \newcommand{\tabincell}[2]{\begin{tabular}{@{}#1@{}}#2\end{tabular}}
	\centering
	\caption{Results of Mask R-CNN and its quantized networks by our method on the COCO validation set.}
	\setlength{\tabcolsep}{1.7mm}{
		\renewcommand\arraystretch{1.2}
		\begin{tabular}{c|c|cccccc|cccccc}
			\hline
			 \multirow{2}*{Methods}  & \multirow{2}*{W/A-Bits} & \multicolumn{6}{c|}{Detection} & \multicolumn{6}{c}{Segmentation} \\ \cline{3-14}
            && AP    & $\textrm{AP}_{50}$    & $\textrm{AP}_{75}$    & $\textrm{AP}_{S}$ & $\textrm{AP}_{M}$ & $\textrm{AP}_{L}$ & AP    & $\textrm{AP}_{50}$    & $\textrm{AP}_{75}$    & $\textrm{AP}_{S}$ & $\textrm{AP}_{M}$ & $\textrm{AP}_{L}$ \\
            \hline
			baseline      & 32/32     & 40.98   & 61.53   & 44.91   & 24.87  & 43.87  & 55.33  & 37.17   & 58.60   & 39.88   & 18.63  & 39.49  & 53.30     \\ \cline{1-14}
			Ours [Haar]   & 32/32     & 41.27(+0.29)   & 61.69   & 45.25   & 25.05  & 44.10  & 53.50 & 37.42(+0.25)   & 58.77   & 40.33   & 18.90  & 39.75  & 53.50     \\
			Ours [db2]    & 32/32     & 41.31(+0.33)   & \textbf{61.87}   & 45.15   & \textbf{25.30}  & 44.27  & 53.50  & \textbf{37.50(+0.33)}   & \textbf{58.97}   & \textbf{40.37}   & \textbf{18.98}  & 39.85  & 53.64    \\
			Ours [sym2]   & 32/32     & 41.30(+0.32)   & 61.80   & 45.10   & 25.08  & 44.14  & \textbf{53.88} & \textbf{37.50(+0.33)}   & 58.88   & 40.29   & 18.87  & 39.78  & 53.69     \\
            Ours [coif2]  & 32/32     & \textbf{41.33(+0.35)}   & 61.79   & \textbf{45.43}   & 25.16  & \textbf{44.28}  & 53.53 & 37.45(+0.28)   & 58.96   & 40.33   & 18.93  & \textbf{39.89}  & \textbf{53.79}    \\ \cline{1-14}
            baseline      & 4/4     & 39.22   & 59.70   & 42.50   & 23.30  & 42.28  & 50.84  & 35.36   & 56.62   & 37.68   & 17.35  & 37.69  & 50.58     \\ \cline{1-14}
			Ours [Haar]   & 4/4     & 39.98(+0.76)   & \textbf{60.43}  & 43.58   & \textbf{24.57}  & 42.83  & 52.08 & \textbf{36.23(+0.87)}   & \textbf{57.55}   & \textbf{38.88}   & \textbf{18.37}  & \textbf{38.46}  & 52.25     \\
			Ours [db2]    & 4/4     & 39.94(+0.72)   & 60.42   & 43.36   & 23.65  & 42.76  & 52.26  & 36.14(+0.78)   & 57.51   & 38.65   & 17.89  & 38.28  & \textbf{52.26}    \\
			Ours [sym2]   & 4/4     & 39.96(+0.74)   & 60.38   & 43.54   & 23.48  & \textbf{42.86}  & \textbf{52.55} & 36.02(+0.66)   & 57.32   & 38.61   & 17.67  & 38.17  & 51.78     \\
            Ours [coif2]  & 4/4     & \textbf{40.06(+0.84)}   & 60.38   & \textbf{43.71}   & 23.74  & 42.81  & 52.32 & \textbf{36.23(+0.87)}   & 57.33   & 38.83   & 17.87  & 38.32  & 52.21      \\
			\hline
		\end{tabular}
	}
	\label{tab:Resnettable}
\vspace{-0.1cm}
\end{table*}

\textbf{Experimental results:}
We evaluate the proposed information enhancement method for object detection and instance segmentation tasks on the COCO detection benchmark \cite{lin2014microsoft},
which is one of the most popular large-scale benchmark datasets.
This dataset consists of images from 80 different categories.
Here, we use the 115K images for training and 5K images for validation.
We use ResNet-50 as the backbone of Mask R-CNN \cite{he2017mask}, in which the feature maps are decomposed for information enhancement.
\emph{Haar}, \emph{Daubechies}, \emph{Symlets}, and \emph{Coiflets} are applied as wavelet bases to decompose the feature maps.
We do experiments on full-precision networks and 4-bit quantized networks.
Here, we use Eq.\ (11) as the quantizer to quantize the weights ${\mathbf{W}}$ and enhanced feature maps $\hat{\mathbf{X}}$.
In the implementation, we apply the pre-trained full-precision model to initialize our model.
Our network is fine-tuned with SGD for 100K iterations with the initial learning rate being $1 \times 10^{-3}$ and the batch size of 16 for 8 V100 GPUs.
The learning rate is decayed by a factor of 10 at iterations 20K, 50K, and 80K, respectively. The initial scale factors $\alpha ^i$ are set to $1.2$.

\begin{figure}[t]
    \setlength{\abovecaptionskip}{0.1cm}
    \setlength{\belowcaptionskip}{-0.3cm}
	\centering
	\includegraphics[width=3.2in]{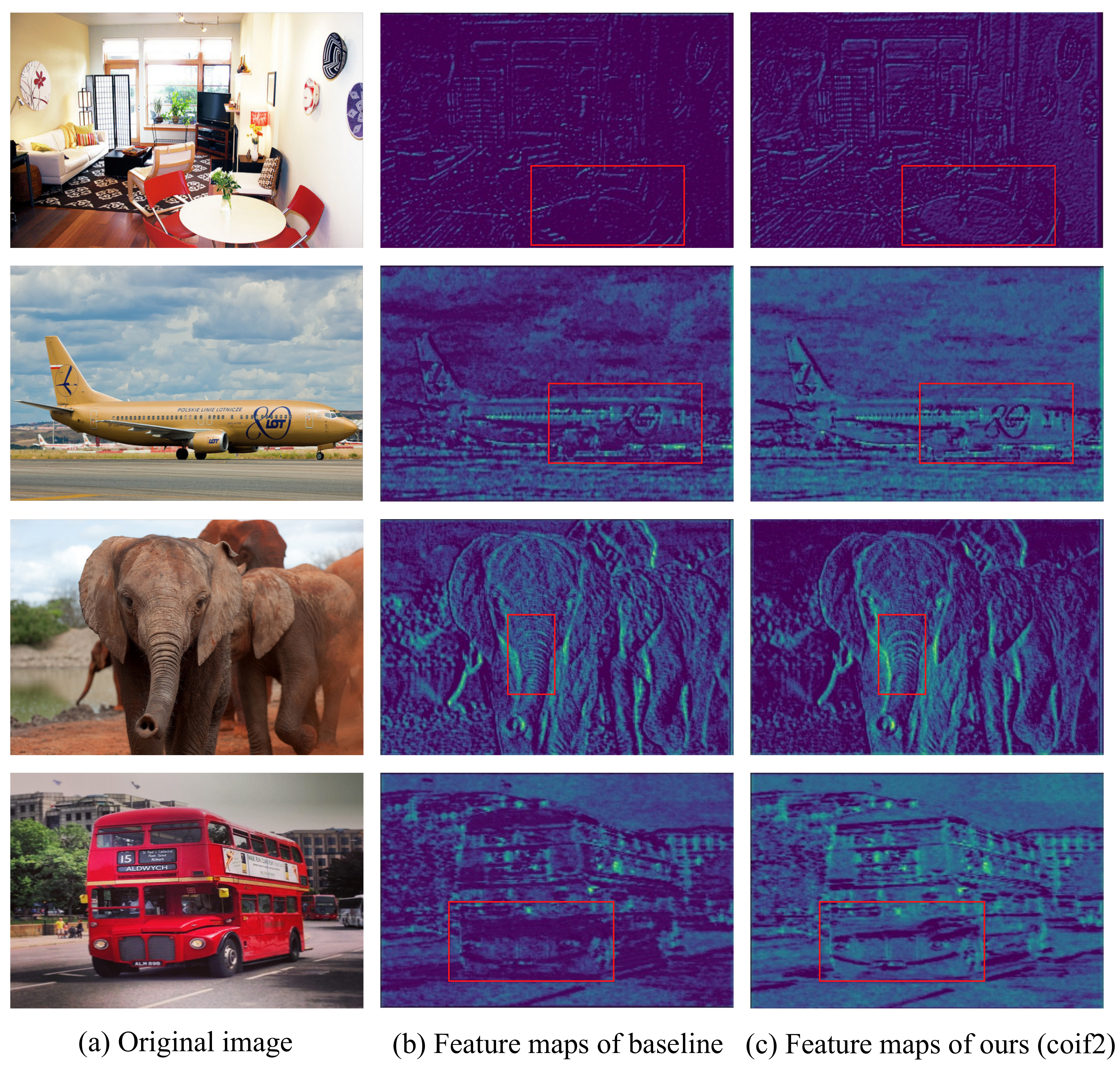}
	\caption{\small{Feature visualization of ResNet-50 in Mask R-CNN.
    }}
\end{figure}

We report the standard COCO metrics including $\textrm{AP}$, $\textrm{AP}_{50}$, $\textrm{AP}_{75}$, $\textrm{AP}_{S}$, $\textrm{AP}_{M}$, and $\textrm{AP}_{L}$ in object detection and instance segmentation tasks.
Experimental results are shown in Table 2.
The results have been significantly improved after enhancing the high frequency information.
No matter what kind of wavelet decomposition base is used,
the average over IOU thresholds of the full-precision model for detection and segmentation tasks have been improved by more than 0.3 (compared with baseline).
For the quantized network, the effect of information enhancement is particularly significant.
As can be seen from Table 2, the enhanced networks have significant improvements
in terms of the metrics for both tasks (e.g., object detection and instance segmentation) compared with their baselines.
Fig.\ 9 shows the visualization of the feature maps (`res2') of quantized backbone (ResNet-50),
where (b) shows original feature maps and (c) shows enhanced feature maps by `coif2'.
The enhancement effect can be clearly seen in the red box.

\vspace{-1 mm}
\subsection{Effectiveness Analysis}
\vspace{-1 mm}

MWQ contains more operations, and thus it will inevitably require more computation, as shown in Fig.\ 2 .
In order to avoid the influence of complex computation on the efficiency of quantized neural networks,
we only introduce MWQ in the training process for Applications 1 and 2.
Although the compressed model needs only once additional IDWT reconstruction to participate in inference computing,
the computation of IDWT is negligible in practical applications.
Experiment 3, as an extended application, verified the importance of spatial information (high frequency components), especially in quantized neural networks.
Although the multiscale frequency and spatial information of the weights is not interpretable as the activation feature maps,
and it also can play an effective role in quantized neural networks.


\vspace{-1 mm}
\section{Concluding Remarks}
\vspace{-1 mm}

In this paper, we proposed a novel quantization method, which considers the spatial information and multiscale information decomposed by wavelet transform.
It can alleviate the information loss by matching the appropriate quantization for each frequency component.
We verified the effectiveness of the proposed method on model compression, quantized network optimization and information enhancement applications.
Due to the flexibility of MWQ,
the discussion in this paper may not be complete, and there are still many innovations and applications
(e.g., speech recognition, denoising, and remote sensing image processing) waiting to be explored.
In the future, we will use MWQ to train a quantized model that supports multiple bit-widths simultaneously to use in more industrial applications.
{\small
\bibliographystyle{ieee_fullname}
\bibliography{egbib}
}

\end{document}